\documentclass[mlmain]{jmlr}

\jmlrproceedings{PMLR}{Proceedings of Machine Learning Research}

\usepackage{booktabs}

\jmlrvolume{}
\firstpageno{1}
\editors{}
\jmlryear{2026}
\jmlrworkshop{Symmetry and Geometry in Neural Representations}

\title[CorrFlow]{CorrFlow: Exact and Permutation-Equivariant Flow Matching on
the Correlation Elliptope}

\author{\Name{Gautier Marti}\Email{gautier.marti@adia.ae}\\
\addr Abu Dhabi Investment Authority (ADIA), Abu Dhabi, United Arab Emirates}

\begin{document}
\maketitle

\begin{abstract}
Correlation matrices combine nonlinear positive-definiteness constraints with
a simultaneous row--column permutation symmetry. Their generation therefore
involves two distinct problems. First, every output must remain on the
correlation elliptope, irrespective of how its variables are ordered. Second,
the generator must learn the domain-specific distribution occupied by
empirical data inside that valid space. Existing financial image
generators mix these problems: they learn ordered matrix entries and repair
invalid draws afterward. We introduce CorrFlow, which separates them. A
generalized Fisher transform maps full-rank correlation matrices bijectively to
unconstrained coordinates, guaranteeing valid outputs without projection. A
graph flow-matching network learns the empirical distribution in those
coordinates and is permutation equivariant: reordering the variables only
reorders the output. We compare a conditional CorrGAN reimplementation, image diffusion, and
CorrFlow on $80\times80$ Taiwan equity matrices using five seeds per neural
method. CorrFlow produces 100\% valid draws without repair and obtains feature
sliced-Wasserstein distance $.712\pm.141$, versus $4.339\pm.252$ for CorrGAN
and $4.610\pm.191$ for image diffusion after their invalid outputs are
projected back to the correlation space. It also improves this distance over
historical lookup, while lookup retains higher local coverage. These results support the combined construction under the evaluated training
budget; they do not establish superiority over fully tuned image models or
isolate the contribution of each component.
\end{abstract}

\begin{keywords}
Correlation matrices, flow matching, permutation equivariance, geometric deep
learning, Riemannian geometry
\end{keywords}

\section{Introduction}

Correlation matrices are structured objects rather than arbitrary arrays. A
full-rank correlation matrix lies in the elliptope's interior: it is symmetric,
positive definite, and has unit diagonal. Relabeling its variables acts by
simultaneously permuting rows and columns, $C\mapsto PCP^\top$, without changing
the underlying pairwise relationships. A generative model should therefore
respect both the geometry of its sample space and this permutation symmetry.

Finance provides a demanding instance of this problem. Let $C$ be an asset
correlation matrix and $D$ the diagonal matrix of asset volatilities. Their
covariance matrix is $\Sigma=DCD$, so a portfolio with weights $w$ has variance
$w^\top\Sigma w$. Changing $C$ can alter diversification, risk-constrained
leverage, and potential losses. Libraries of plausible matrices support
correlation stress testing,
reverse stress testing, and portfolio-risk analysis
\citep{packham2023scenarios}. Covariance structure is equally central to signal
processing tasks such as adaptive beamforming and source detection
\citep{aubry2021radar}. We focus on finance, where crisis observations are scarce
and the number of pairwise correlations grows quadratically with the asset
universe.

CorrGAN \citep{marti2020corrgan} showed that a GAN can recover empirical
spectral, hierarchical, and network properties missed by generic random-matrix
samplers; cCorrGAN added market-regime conditioning
\citep{marti2021ccorrgan}. A later image DDPM study reported stable training and
better volatility prediction in a case study \citep{kubiak2024ddpm}. These methods first use
hierarchical clustering to turn an unordered matrix into an image. Their output
depends on that ordering and may leave the correlation space, requiring a
nearest-correlation repair.

Manifold-aware diffusion is possible \citep{li2024spdddpm}, while DiffeoCFM
showed that a global matrix diffeomorphism can turn Riemannian conditional flow
matching into ordinary Euclidean flow matching \citep{collas2025diffeocfm}. Its
normalized Cholesky coordinates, however, remain order dependent. CorrFlow
uses the same pullback principle with GFT coordinates and an equivariant vector
field; it does not introduce a new general framework for Riemannian flow
matching. A normalized-Cholesky control and an objective comparison are
discussed in Appendix~\ref{app:ablations}; an additional five-seed comparison
with the adapted DiffeoCFM and its fitted Gaussian prior appears in
Section~\ref{sec:matchedcategorical}.

CorrFlow assigns the constraints and distribution learning to separate
components. The generalized Fisher transform (GFT) converts a full-rank
correlation matrix into an unconstrained vector, and its inverse converts every
finite vector back into a valid correlation matrix. CorrFlow learns the
distribution of these vectors with a permutation-equivariant graph network. In
plain terms, reordering the variables only reorders the generated matrix.

This paper makes two contributions. First, it combines an exact correlation
parametrization with a provably permutation-equivariant flow. Second, it gives
a controlled Taiwan case study against conditional versions of CorrGAN and an
image DDPM, a low-rank Gaussian generator, and historical lookup. All neural
models see the same matrices and conditions, and neural results in the primary
comparison use five seeds. The primary Taiwan training and test windows share
no return sessions; adjacent split boundaries remain overlapping. The purpose is to evaluate a
generative model of empirical correlation matrices, not to forecast returns or
the next correlation matrix.

\begin{figure}[t]
\centering
\includegraphics[width=.82\textwidth]{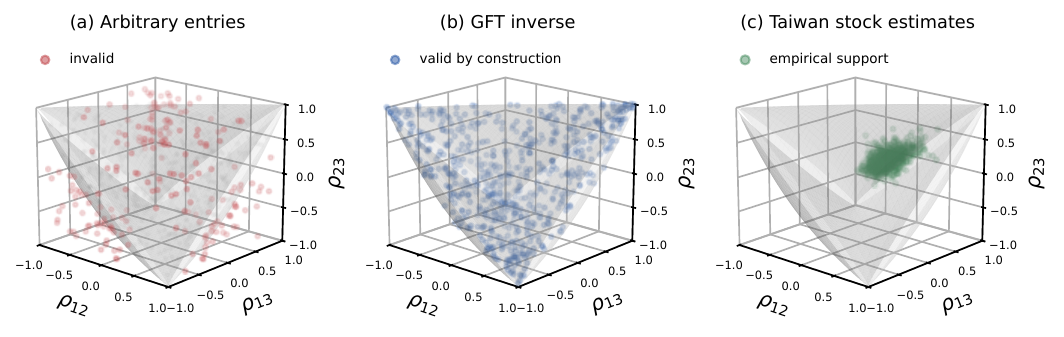}
\caption{Geometry versus domain distribution for $3\times3$ matrices. Valid
correlations lie inside the gray elliptope
($1+2\rho_{12}\rho_{13}\rho_{23}-\rho_{12}^2-\rho_{13}^2-\rho_{23}^2>0$).
(a) Choosing entries independently produces invalid points. (b) Every
unconstrained GFT vector maps inside the elliptope. (c) Nine hundred random
three-stock submatrices from the Taiwan training windows occupy a much smaller,
mostly positive region. GFT enforces the universal constraint; CorrFlow learns
the domain-specific distribution within it.}
\label{fig:elliptope}
\end{figure}

\section{CorrFlow}

\subsection{Generalized Fisher coordinates}

\textbf{Intuition.} We first remove the constraints by a reversible change of
coordinates. Let
$\mathcal E_n^+=\{C\in\mathbb S_{++}^n:\operatorname{diag}(C)=\mathbf 1\}$.
This is the interior of the correlation elliptope and precisely the set called
``valid'' here: symmetric, positive-definite, unit-diagonal matrices. The
generalized Fisher transform
(GFT)
\begin{equation}
  \gamma(C)=\operatorname{vecl}(\log C)\in\mathbb R^{n(n-1)/2}
\end{equation}
stacks the entries strictly below the diagonal of $\log C$. It is a bijection
from $\mathcal E_n^+$ to an unconstrained Euclidean space
\citep{archakov2021parametrization}. Its inverse recovers the missing diagonal of
$\log C$ by fixed-point iteration before exponentiation. Consequently, every
finite generated vector maps to exactly one valid $C$, with no clipping or
nearest-correlation projection. Figure~\ref{fig:elliptope} visualizes the
geometry of the correlation elliptope for $3\times3$ matrices, the smallest
matrix size in which positive definiteness jointly constrains multiple
correlation coefficients.

The Euclidean inner product in GFT coordinates induces the pullback metric
$g_C(U,V)=\langle D\gamma_C[U],D\gamma_C[V]\rangle$ on $\mathcal E_n^+$. In
plain terms, Euclidean distances and paths after transformation define a
Riemannian geometry on the original elliptope. CorrFlow can therefore be
trained using ordinary Euclidean flow matching in GFT coordinates; it does not
require the explicit computation of manifold geodesics.

To express the permutation symmetry, write $\Gamma(C)$ for the symmetric,
zero-diagonal (``hollow'') matrix whose lower triangle is $\gamma(C)$. Thus
$\gamma(C)$ is a vector, while $\Gamma(C)$ arranges the same coordinates as a
matrix. Unlike
order-dependent Cholesky coordinates, this representation is permutation
equivariant. For any permutation matrix $P$,
\begin{equation}
  \Gamma(PCP^\top)=P\Gamma(C)P^\top.                 \label{eq:gft-equiv}
\end{equation}
Thus, relabeling the variables before transformation merely applies the
corresponding row--column relabeling to the GFT representation; it does not
introduce a privileged ordering.

\subsection{Equivariant rectified flow}

\textbf{Flow-matching objective.} Let $\mu$ and $\sigma$ be the scalar mean and
standard deviation of all GFT edges in the training set. Define
$S(X)_{ij}=(X_{ij}-\mu)/\sigma$ for $i\ne j$ and $S(X)_{ii}=0$. We train on the
standardized endpoint $X_1=S(\Gamma(C))$; using the same two scalars on every
edge preserves permutation equivariance. Let $X_0$ be a hollow symmetric matrix
whose distinct edges are independent standard Gaussian variables. For
$t\sim\operatorname{Uniform}(0,1)$, define the straight interpolation between
the sampled pair. Rectified flow trains a vector field to recover its constant
velocity \citep{liu2023rectified}:
\begin{equation}
 \begin{aligned}
 \mathcal L(\theta)&=\mathbb E\left\|v_\theta(X_t,t,y)-(X_1-X_0)\right\|_F^2,\\
 X_t&=(1-t)X_0+tX_1.
 \end{aligned}                                      \label{eq:rf}
\end{equation}
Here $y$ denotes global conditioning information that is unchanged by a
permutation of the variables. In our financial experiment, $y$ consists of the
regime indicator and continuous performance score defined in
Section~\ref{sec:experiments}.

At generation time, the learned vector field defines the ODE
\begin{equation}
 \frac{\mathrm dX(t)}{\mathrm dt}=v_\theta(X(t),t,y),\qquad X(0)=X_0,
 \label{eq:flow-ode}
\end{equation}
and we denote its endpoint by $F_\theta(X_0,y)=X(1)$. Thus solving the ODE
transports fresh Gaussian noise to a standardized GFT sample
\citep{lipman2023flow}. Applying $S^{-1}$ and then the inverse GFT produces the
generated correlation matrix.

\textbf{Equivariant vector field.} We represent each off-diagonal matrix entry
as an edge of a complete graph. Each layer first updates node $i$ from the mean
and second moment of its incident edge states, the graph-wide edge mean, and a
context embedding of $(t,y)$. It then updates edge $(i,j)$ with one shared MLP
applied to its previous state, $h_i+h_j$, $h_i\odot h_j$,
$|h_i-h_j|$, and the context. Residual connections and layer normalization are
used throughout. Because the same functions are applied at every node and edge,
the aggregations are invariant, and the endpoint combinations are symmetric,
the vector field satisfies
\begin{equation}
 v_\theta(PXP^\top,t,y)=Pv_\theta(X,t,y)P^\top.       \label{eq:v-equiv}
\end{equation}
Mean aggregation normalizes for graph degree, so its scale does not grow
directly with the matrix dimension. The same parameters can therefore be
evaluated at different dimensions, although this alone guarantees neither
cross-size accuracy nor projective consistency. For fixed width and depth, each vector-field evaluation uses $O(n^2)$
computation and memory. This is not the cost of the complete sampler: with
$K$ evaluations and $I$ inverse-GFT iterations, dense linear algebra gives
$O(Kn^2+In^3)$ time. The matrix logarithm used in preprocessing also costs
$O(n^3)$. Appendix~\ref{app:runtime} separates recorded training and sampling
costs.

\textbf{Guarantee.} Equations~\eqref{eq:gft-equiv} and~\eqref{eq:v-equiv}, the
fact that $S$ commutes with permutations, and ODE uniqueness give the following
result.

\begin{figure}[t]
\centering
\includegraphics[width=.96\textwidth]{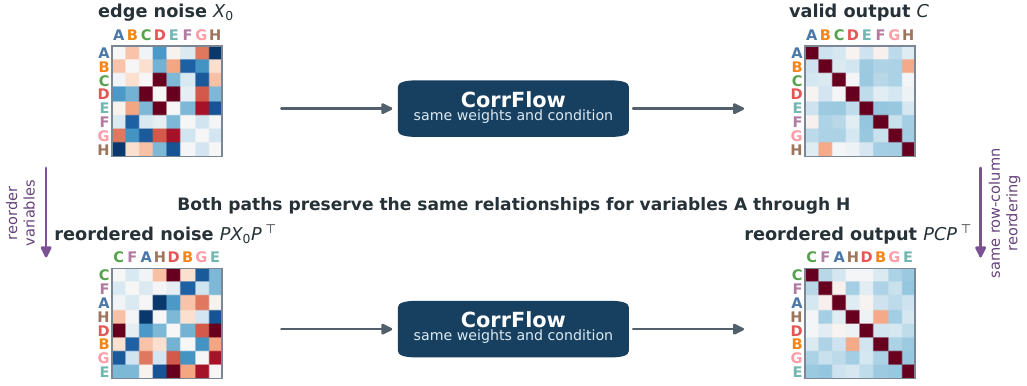}
\caption{What permutation equivariance guarantees. Colors track eight variable
identities. The top path maps edge noise $X_0$ to a valid correlation matrix
$C$. The bottom path applies the same permutation $P$ to the input and uses the
same CorrFlow weights and condition; its output is exactly the corresponding
row--column permutation $PCP^\top$. Thus relationships follow their variables
and the model does not learn a privileged ordering. The two paths agree exactly
in exact arithmetic; a numerical implementation may differ only through
floating-point rounding.}
\label{fig:equivariance}
\end{figure}

\begin{proposition}
For fixed $y$, suppose that $v_\theta$ satisfies~\eqref{eq:v-equiv} and that
the ODE has a unique solution on $[0,1]$ for each initial condition under
consideration. Its flow map then obeys
$F_\theta(PX_0P^\top,y)=PF_\theta(X_0,y)P^\top$. Moreover, every finite endpoint
produces a matrix in $\mathcal E_n^+$ after unstandardization and inverse GFT.
\end{proposition}
\noindent The first claim follows because permuting a solution gives the unique
solution starting from the permuted initial condition. The second follows from
the GFT bijection. Thus changing row--column labels changes only the labels of
the output. Because the Gaussian base distribution is also permutation
invariant, the resulting distribution of correlation matrices is exchangeable
for the anonymous-asset model used here \citep{klein2023equivariant}.
Figure~\ref{fig:equivariance} illustrates the same statement without notation.

\section{Experimental Protocol}
\label{sec:experiments}

\textbf{Data.} We construct the primary benchmark from curated daily returns
for TWSE/TPEx stocks, retaining window dates and constituent identifiers.
Appendix~\ref{app:corrgan} details the return calculation. For
every eligible session we draw eight random 80-asset subsets from the 500 most
liquid stocks then available and estimate correlation over the preceding 252
sessions. This gives 24,408 matrices: 13,600 for training (2014--2020), 3,920
for validation (2021--2022), and 6,888 for testing (2023--July 2026).
Partitions are chronological by window endpoint, but windows across adjacent
boundaries still share underlying return observations.
Development used other
asset subsets from the same test dates; the final resampling excludes duplicate
date/asset sets but is not an untouched temporal holdout.

All methods use these same return-derived matrices and chronological splits.
We retain $n=80$ for comparability with earlier cCorrGAN experiments and because it
is well below the rank ceiling of 251 for 252 demeaned observations. This is a
benchmark choice, not a model-size limit. Larger dimensions require separate
conditioning, runtime, and fidelity checks (Appendix~\ref{app:size}). Taiwan
provides dated, identified constituents for chronological evaluation; this
primary experiment does not establish transfer to the S\&P 500.
Appendix~\ref{app:japan} reports a separate five-seed Japan diagnostic.

For each window, define $q=\sqrt{252}\,\bar r/s_r$ as a continuous performance
score, where $\bar r$ and $s_r$ are the mean and standard deviation of the
equal-weight portfolio return in that window. The bottom 20\%,
middle 60\%, and top 20\% of training scores define adverse, typical, and
favorable regimes. A condition consists of the regime indicator and the
standardized continuous score. It lets the model associate different
correlation structures with broad observed market states; no part of the
experiment predicts $q$.

\textbf{Baselines.} Both replay controls draw stored training matrices from the
requested regime. Historical replay samples uniformly; matched replay samples
from the 100 training windows with the closest performance scores. They are
strong lookup controls, not learned generators.
Gaussian models provide simple non-neural baselines. For elliptical data,
\citet{archakov2026gft} report approximately Gaussian, weakly correlated GFT
estimation errors; this does not imply a Gaussian law across market windows.
PCA-Gaussian is fitted separately within each
regime. It models the matrix collection, not stock returns: within a regime, it
fits a Gaussian to the leading 64 principal components of the
3,160-dimensional GFT vectors and adds isotropic residual noise. Matched PCA
also lets the mean vary linearly with the continuous performance score.

The two image baselines receive the same regime and continuous score as
CorrFlow. Conditional CorrGAN is our PyTorch implementation of the published
cCorrGAN architecture and adversarial objective \citep{marti2021ccorrgan},
extended with the continuous performance score.
Image DDPM uses a
convolutional U-Net, a 1,000-step cosine noise schedule, and 50-step DDIM
sampling \citep{song2021ddim}. Following CorrGAN, each empirical matrix is first put into a
canonical-looking image order by Ward clustering with optimal leaf ordering.
Because neither image model guarantees a correlation matrix, we copy the upper
triangle, set the diagonal to one, and repair it using Higham's
nearest-correlation procedure followed by a small adjustment for strict positive
definiteness \citep{higham2002nearest} before measuring fidelity. The revised
image scores use float64 repair to tolerance $10^{-6}$; all runs converge
within the 1,000-iteration cap. Appendix~\ref{app:corrgan} records the correction. ``Validity before repair''
means positive definiteness after the first two deterministic steps but before
the projection. CorrFlow sees the matrices in their original random order and
requires neither ordering nor repair. Appendix~\ref{app:corrgan} distinguishes
this retrained baseline from a separate check of an author-released CorrGAN
checkpoint. Reproduction of the published S\&P fidelity results by the
conditional reimplementation remains unestablished.

\textbf{Metrics.} Let $s(C)$ contain 21 order-invariant summaries: correlation
and eigenvalue quantiles, the largest and top-five eigenvalues, two effective
ranks, log determinant, and log condition number. We scale each summary by its
training standard deviation; Appendix~\ref{app:summaries} lists all 21.
Feature sliced-Wasserstein distance (F-SWD) averages one-dimensional $W_2$
distance over 128 random projections \citep{bonneel2015sliced}; energy distance
is a complementary multivariate comparison \citep{szekely2013energy}. Lower is
better. Using five-nearest-neighbor neighborhoods, manifold precision asks how
often generated matrices resemble held-out matrices, while recall asks how much
of the held-out variety the generator reproduces
\citep{kynkaanniemi2019precision}; higher is better. Validity before repair uses
tolerance $10^{-5}$.

All methods are compared with the same 1,500 test cases, 500 from each regime.
For each case, a conditional method receives its regime and continuous score.
The pooled metrics therefore compare conditional fit under this balanced set,
not under the test period's natural regime frequencies. We also train a logistic
classifier to distinguish real from generated matrices from the same 21
summaries \citep{lopezpaz2017c2st}. Its five-fold cross-validated area under the
receiver operating characteristic curve (AUC) is .5 at chance and approaches 1
as the difference becomes easier to detect. Every AUC uses 20 repeated samples
of 250 matrices per regime on each side. Finally, real--real resampling splits
genuine test data in two and supplies the finite-sample reference: no fitted
model is expected to reach zero. A nearest-training-distance check separately
diagnoses memorization.

\textbf{Training.} Every neural method is trained for 20 passes over the same
13,600 matrices using seeds 1701--1705. CorrFlow has 48 hidden channels, three
edge--node blocks, 0.17M parameters, batch size 64, and AdamW at
$3\times10^{-4}$. Its GFT edge coordinates use the shared-scalar standardization
$S$ defined above. Generation uses 25 Heun steps
(50 vector-field evaluations), where Heun's method is a second-order
predictor--corrector ODE solver. CorrGAN has 5.56M generator-plus-discriminator parameters and
uses batch size 64 and Adam at $2\times10^{-4}$; DDPM has 3.84M parameters and
uses batch size 32 and AdamW at $2\times10^{-4}$. CorrGAN is evaluated at the fixed final epoch. CorrFlow
and DDPM select the raw minimum-validation-loss checkpoint. Neural models use
five training seeds; replay and PCA controls use five generation seeds on the
same held-out cases. Table entries report means and sample standard deviations. Equal epochs do
not equalize optimization, compute, or convergence; longer validation-selected
image-model training could change the comparison.
The public software is organized as
\href{https://github.com/gautiermarti/CorrGAN}{\texttt{CorrGAN}} (CorrGAN/cCorrGAN),
\href{https://github.com/gautiermarti/CorrFlow}{\texttt{CorrFlow}} (geometric generators),
and \href{https://github.com/gautiermarti/CorrBench}{\texttt{CorrBench}} (return-based
construction and shared evaluation), with coordinated version \texttt{0.1.0rc2}.
Both model repositories pin the compatible CorrBench revision.
The underlying prices are available through official public TWSE and
TPEx endpoints. Experiments ran on an NVIDIA DGX Spark with GB10.

\section{Results}

\begin{table}[t]
\caption{Taiwan test results. Distances ($\downarrow$) use repaired image
outputs; validity before repair, precision, and recall ($\uparrow$) are better
when higher. For classifier AUC, .5 means indistinguishable and 1 perfectly
separable. Method rows give mean with sample s.d. in parentheses over five
training or generation seeds; AUC is first averaged over 20 matched subsamples
within each seed. Real-data references use 20 resamples. Bold marks the best
fitted generator. Matched rows use the continuous performance score.}
\label{tab:final}
\centering
\scriptsize
\setlength{\tabcolsep}{4pt}
\resizebox{\textwidth}{!}{%
\begin{tabular}{lrrrrrr}
\toprule
Model & Valid before repair$\uparrow$ & F-SWD$\downarrow$ & Energy$\downarrow$ & Precision$\uparrow$ & Recall$\uparrow$ & Classifier AUC\\
\midrule
\multicolumn{7}{l}{\emph{Real-data references}}\\
Real--real & -- & .061 (.009) & .010 (.003) & .890 (.012) & .896 (.014) & .499 (.021)\\
Train--test temporal shift & -- & -- & -- & -- & -- & .950 (.005)\\
\midrule
\multicolumn{7}{l}{\emph{Historical lookup (not generative)}}\\
Historical replay & 1.000 (.000) & .886 (.010) & 2.715 (.072) & .457 (.011) & .589 (.036) & .951 (.003)\\
Matched replay & 1.000 (.000) & .899 (.011) & 2.820 (.057) & .424 (.006) & .576 (.024) & .950 (.003)\\
\midrule
\multicolumn{7}{l}{\emph{Non-neural fitted generators}}\\
PCA-64 Gaussian & 1.000 (.000) & .953 (.009) & 3.634 (.054) & .168 (.011) & .133 (.035) & .9998 (.00009)\\
Matched PCA & 1.000 (.000) & .965 (.023) & 3.694 (.149) & .162 (.007) & .114 (.051) & .9998 (.00004)\\
\midrule
\multicolumn{7}{l}{\emph{Neural fitted generators}}\\
cCorrGAN (reimplementation) & .000 (.001) & 4.339 (.252) & 39.984 (1.958) & .000 (.000) & .200 (.447) & 1.0000 (.00000)\\
Image DDPM & .001 (.001) & 4.610 (.191) & 36.617 (2.139) & .000 (.000) & .200 (.447) & 1.0000 (.00000)\\
CorrFlow & \textbf{1.000 (.000)} & \textbf{.712 (.141)} &
\textbf{1.885 (.722)} & \textbf{.393 (.090)} & \textbf{.435 (.082)} &
\textbf{.9981 (.00026)}\\
\bottomrule
\end{tabular}
}
\end{table}

Within Table~\ref{tab:final}, CorrFlow is the only neural method valid before
repair and has the best mean fidelity and coverage metrics among fitted
generators under the reported budget. Projection
makes every image output numerically valid. Measured as root-mean-square change
per off-diagonal correlation, the repair is $.0179\pm.0049$ for CorrGAN and
$.0175\pm.0029$ for DDPM across seeds.
Neither image architecture enforces the sample-space constraints. Their weak
fidelity relative to PCA also warrants a stronger training-budget check; raw
invalidity alone does not explain the post-repair fidelity gap. After repair, both image models are
still easily separated from real data (classifier AUC 1.000 in every seed).

CorrFlow also improves F-SWD and energy distance over historical lookup. This
does not make lookup obsolete: replay has higher local precision and recall,
and retains observed scenarios exactly. Conversely, none of the 7,500 CorrFlow
draws duplicates a training GFT vector. Its median distance to the nearest
training example is $.845\pm.101$ times the corresponding real-test distance,
whereas replay's ratio is $.513\pm.004$.

Classifier results require a temporal reference. Two stratified subsets of the
test period are indistinguishable (AUC $.499\pm.021$), but genuine training- and
test-period matrices are strongly separable (AUC $.950\pm.005$). Figure~
\ref{fig:spectral-shift} shows that this shift is strongly spectral. Test
matrices place more variance in their leading eigenvalues and have fewer
effective spectral dimensions. Because a correlation matrix has fixed trace
$n$, this means that a few common modes explain more of the standardized
variance. For broad portfolios with comparable marginal volatilities, it
suggests stronger common movement and less diversification, although realized
risk also depends on volatilities and weights.

Historical replay consequently remains detectable at AUC $.951\pm.003$.
CorrFlow is more detectable at AUC $.9981\pm.00026$, so temporal shift does not
explain the entire discrepancy: the generated and real summary distributions remain systematically
different despite the smaller aggregate distances. A linear classifier on the
summaries does not isolate errors in their joint dependence from marginal
mismatches. Lower F-SWD and energy therefore do not establish recovery of the
full held-out variety or joint law. The claim is therefore narrower: among fitted
generators, CorrFlow combines exact validity, order independence, novelty, and
better aggregate fidelity within this finite-budget comparison.

\section{What the Guarantees Solve and What They Do Not}

The comparison motivates studying representation separately from the
adversarial, diffusion, and flow objectives; Table~\ref{tab:final} changes both
and cannot assign their individual causal contributions. An image DDPM still
requires projection because its denoising process lives in unconstrained pixel
space. CorrFlow is useful not because flow matching is universally superior to
diffusion, but because its learned dynamics live in exact, equivariant
coordinates. The strong PCA control likewise prevents attributing every gain to
neural capacity. Appendix~\ref{app:ablations} reports exploratory coordinate,
architecture, and matched-objective diagnostics with their budget limitations.

The target is the distribution of \emph{estimated} 252-day correlations, which
mixes latent dependence, estimator noise, and heavily overlapping windows. The
24,408 Taiwan matrices are therefore not 24,408 independent market episodes.
The performance score is conditioned upon, not predicted. Total exchangeability suits anonymous
random subsets, but named-asset simulation should carry sector, country,
liquidity, and factor attributes along with each row and column. Correlation
alone also omits volatility, marginal tails, and dynamics required for a full
market-risk simulator.

The next mathematical requirement is projective consistency: generating $k$
assets directly should have the same law as taking a random $k\times k$
principal submatrix from a larger draw. It is not implied by permutation
equivariance. Multi-size training, node attributes, and separating the
underlying dependence structure from the sampling noise introduced by finite
return windows are important extensions, but belong to a broader study rather
than to the present claim. Alternative exchangeable source laws, such as
heavy-tailed edge noise or a regime-conditioned mixture, could change transport
difficulty while preserving equivariance; we have not evaluated them.

\section{Conclusion}

CorrFlow replaces image heuristics and post-hoc repair with exact elliptope
coordinates and a permutation-equivariant flow. This combination guarantees
valid outputs and removes arbitrary asset order from the learned mechanism. It
generates new rather than retrieved scenarios and, in the Taiwan case study,
improves aggregate structural distances over our 20-epoch cCorrGAN and image
DDPM implementations and historical lookup. Coverage remains incomplete.
CorrFlow provides a geometric framework for learning distributions of
correlation matrices.

\acks{
OpenAI Codex was used to assist with code development, experiment
orchestration, figure preparation, and language-editing suggestions throughout
the manuscript. The author designed the study, tested the implementation,
verified the data processing, numerical results, mathematical claims, and
citations, substantially revised the manuscript, and takes full responsibility
for its content.
}

\bibliography{references}

\clearpage
\appendix
\section{Structural Summary Statistics}
\label{app:summaries}

Let $r_{ij}$ denote the distinct off-diagonal correlations, let
$0<\lambda_1\leq\cdots\leq\lambda_n$ be the eigenvalues of $C$, and let
$p_i=\lambda_i/\sum_j\lambda_j=\lambda_i/n$. Table~\ref{tab:summaries}
lists the permutation-invariant summaries used for F-SWD, energy distance,
manifold precision and recall, and classifier AUC.

\begin{table}[h]
\caption{The 21 structural summaries used in distributional evaluation.}
\label{tab:summaries}
\centering
\small
\begin{tabular}{p{.22\textwidth}cp{.29\textwidth}p{.30\textwidth}}
\toprule
Group & Count & Definition & Interpretation\\
\midrule
Correlation moments
& 2
& Mean and standard deviation of $\{r_{ij}:i<j\}$
& Overall correlation level and cross-sectional dispersion.\\
\addlinespace
Correlation quantiles
& 7
& Quantiles $.01,.05,.25,.50,.75,.95,.99$ of $\{r_{ij}:i<j\}$
& Shape and tails of the pairwise-correlation distribution.\\
\addlinespace
Eigenvalue quantiles
& 6
& Quantiles $0,.05,.25,.50,.75,.95$ of $\{\lambda_i\}_{i=1}^n$
& Shape of the bulk spectrum, including its smallest eigenvalue.\\
\addlinespace
Leading eigenvalue mass
& 2
& $\lambda_n$ and $\sum_{i=n-4}^{n}\lambda_i$
& Variance concentrated in the largest common mode and in the five largest
components.\\
\addlinespace
Effective ranks
& 2
& $\exp(-\sum_i p_i\log p_i)$ and $1/\sum_i p_i^2$
& Effective number of independent spectral dimensions; smaller values mean
greater concentration.\\
\addlinespace
Log determinant
& 1
& $\log\det C=\sum_i\log\lambda_i$
& Spectral volume, with strong sensitivity to small eigenvalues.\\
\addlinespace
Log condition number
& 1
& $\log(\lambda_n/\lambda_1)$
& Spectral ill-conditioning and proximity to singularity.\\
\midrule
Total & 21 & &\\
\bottomrule
\end{tabular}
\end{table}

\clearpage
\section{Baseline Provenance and Scope}
\label{app:corrgan}

The Taiwan source field \texttt{daily\_total\_return} primarily stores
\(\Delta/(P-\Delta)\), where \(P\) is the closing price and \(\Delta\) the
exchange-reported signed change. The curator uses this return when the implied
reference price is positive and its magnitude is at most .35; otherwise it
tries the raw change from the previous observed close. Missing candidates or
magnitudes above .35 become null. An arithmetic audit reproduces all 5,534,321
panel rows exactly, including 7,877 finite raw-close fallbacks and 160 nulls.
The internal column name is retained for reproduction, but this convention is
not an independently reconciled cash-dividend total-return series. No event-level
corporate-action or delisting-proceeds audit is implied.

The primary Taiwan comparison starts from dated stock returns: window
selection, correlation estimation, and regime construction are shared across
methods. The legacy S\&P matrix archive and released CorrGAN checkpoint support
separate implementation checks; they do not supply data for the Taiwan table.
Their missing provenance limits the historical checks as described below and
in Appendix~\ref{app:ablations}.

The post-review numerical audit re-evaluates the original image-model draws;
no image model was retrained for Table~\ref{tab:final}. The submitted evaluation
used 100 float32 repair iterations and did not flag nonconvergence as an error.
All samples were positive definite after the strict-PD adjustment, but the
fixed-point tolerance was not met in every batch. Repeating repair in float64
with a 1,000-iteration cap converges for all 15,000 draws. Mean F-SWD changes
from 4.373 to 4.339 for cCorrGAN and from 4.645 to 4.610 for DDPM; the model
ranking is unchanged. Table~\ref{tab:final} and the abstract use these corrected
scores. A longer-budget retraining study is not included in the present table.

The Taiwan cCorrGAN row uses a new PyTorch implementation with an
80-dimensional noise vector, a regime embedding, an added continuous score,
a $20\times20$ projection, two upsampling layers, and a final $7\times7$
convolution. The discriminator receives the same condition. Training uses
binary cross-entropy, Adam with $(\beta_1,\beta_2)=(.5,.999)$, and real-label
smoothing of .1. These implementation and conditioning choices, together with
the fixed 20-epoch budget, distinguish this experiment from a reproduction of
the published S\&P results. Poor performance against PCA is a reason to audit
training and model selection, not evidence that image generators cannot learn
financial correlations.

The author's 2020 CorrGAN release provides pretrained TensorFlow models and
sampling code, including the nearest-correlation
repair step.\footnote{\url{https://marti.ai/qfin/2020/08/11/corrgan-pretrained-models.html}}
Our separate archived audit sampled 1,800 matrices from its unconditional
80-dimensional model using 100-dimensional Gaussian noise. After copying the
upper triangle and setting the diagonal, none was positive definite; the median
minimum eigenvalue was $-1.71$. This confirms the need for the documented repair
in that checkpoint. It neither reproduces the original fidelity evaluation nor
validates the Taiwan conditional reimplementation. In particular, the legacy
checkpoint's training overlap with the cCorrGAN archive is unknown, and its
repair audit did not converge for every sample within its iteration budget.
We therefore do not use its repaired distributional scores as a competitive
baseline. The downloaded model-archive SHA-256 is
\texttt{ed7a0a1ad5025345052ada7b710a87d6c0da}\allowbreak
\texttt{69b087e89867f8d3f005c99b1d68}.

A subsequent native PyTorch import reproduces the author-released unconditional
and categorical generators on 192,000 raw entries each, with maximum errors
\(7.153\times10^{-7}\) and \(2.623\times10^{-6}\) against TensorFlow.
This verifies generator inference, not the original discriminator, optimizer
or training population. We also completed a separate 50-epoch unconditional
reference reconstruction on the legacy archive (9,800 optimizer steps, one
training seed), using explicit reconstructed training choices and validation
only. Minimum native validation SWD selects epoch 45 and reproduces exactly.

On the same 300 uniform validation cases, five generation seeds give structural
SWD \(2.487\pm.015\) for selected weights and \(3.418\pm.083\) for final
weights, versus \(.114\pm.022\) for unconditional training replay. These
diagnostic scores use projection seed 2027; native selection uses seed 44301.
Both learned checkpoints have zero measured structural precision/recall and
two-sample AUC 1.000. All 3,000 raw generated matrices are invalid after
symmetry and unit-diagonal preparation. Converged float64 repair makes every
matrix valid, with eigenvalues near the \(10^{-6}\) floor, but does not resolve
the distribution mismatch. No test data or regime conditions enter this study.
The recovered inference architecture is therefore supported, while this
reference recipe does not reproduce the archive distribution. This failure
does not establish a failure of the published CorrGAN method.

\section{Exploratory Component Diagnostics}
\label{app:ablations}

These archived development experiments are reported separately from the
five-seed Taiwan table. They cannot substitute for a fully matched factorial
ablation. The legacy S\&P archive contains 18,000 matrices, split 70/15/15
within regime; dates and asset identities are unavailable, so this split does
not measure temporal generalization.

\textbf{Coordinates and architecture.} A dense normalized-Cholesky flow uses
\[
 z=\operatorname{vecl}(\operatorname{diag}(L)^{-1}L),
\]
where $L$ is the Cholesky factor of $C$. Its inverse forms a unit-diagonal lower-triangular factor from
$z$, multiplies it by its transpose, and normalizes to a correlation matrix.
This is the correlation coordinate map used by DiffeoCFM
\citep{collas2025diffeocfm}. Our MLP control uses a standard Gaussian source and does not reproduce that
paper's full implementation, which uses fitted class-conditional Gaussian sources. Both GFT and normalized Cholesky guarantee
validity; only GFT commutes with arbitrary simultaneous row--column permutations.

Table~\ref{tab:coordinates} compares these archived flows after recomputing
all inverses in float64 with GFT tolerance $10^{-8}$ and a 2,000-iteration cap.
Every inverse converged (at most 67 iterations) and every matrix is valid.
Relative to the archived float32 matrices, no entry changed by more than
$3.6\times10^{-6}$. The S\&P dense flows share the same approximately
8.5M-parameter MLP, 100-epoch budget, seed 1701, and 600 evaluation cases;
minimum-validation-loss checkpoints occur at epochs 45 (GFT) and 97 (Cholesky).
This offers a single-seed coordinate comparison, not a definitive ranking.
CorrFlow uses 0.17M parameters and 20 epochs. The Taiwan dense flow was trained
for 32 epochs and selected epoch 14; its CorrFlow comparator uses epoch 20.
All samplers use 25 Heun steps. Architecture comparisons therefore also change
capacity and budget, and the Taiwan rows use 300 pilot cases rather than the
final 1,500-case sample holdout.

\begin{table}[h]
\caption{Exploratory coordinate and architecture diagnostics after the numerical
inverse audit; one seed per model. Distances are comparable within each dataset
only. All rows have 100\% validity.}
\label{tab:coordinates}
\centering
\footnotesize
\begin{tabular}{llrrrr}
\toprule
Dataset & Model & F-SWD$\downarrow$ & Energy$\downarrow$ & Precision$\uparrow$ & Recall$\uparrow$\\
\midrule
S\&P & Dense GFT flow & .308 & .430 & .657 & .597\\
S\&P & Dense normalized-Cholesky flow & .335 & .642 & .167 & .157\\
S\&P & CorrFlow & .294 & .393 & .575 & .420\\
\midrule
Taiwan pilot & Dense GFT flow & 1.073 & 4.465 & .347 & .217\\
Taiwan pilot & CorrFlow & .951 & 3.283 & .497 & .603\\
\bottomrule
\end{tabular}
\end{table}

A direct float32 relabeling test on eight S\&P inputs gives mean relative
Frobenius vector-field error $2.15\times10^{-7}$ for CorrFlow and .625 for the
dense GFT flow. This checks architectural equivariance separately from learned
fidelity; it does not establish an isolated fidelity benefit from equivariance.

\textbf{Objective at fixed geometry and architecture.} Table~\ref{tab:objective}
uses the same 169,537-parameter equivariant network, GFT coordinates, Taiwan
training data, conditions, 20 epochs, training seed 1701, 50 network evaluations,
and 1,500 final test cases. Raw weights are selected by each objective's minimum
validation loss. Velocity prediction was tried after observing poor
noise-prediction results and is explicitly exploratory. All inverses converged
and all draws were valid. These single-seed results do not rank flow matching
against the wider family of diffusion models.

\begin{table}[h]
\caption{Exploratory objective comparison; one matched training seed.}
\label{tab:objective}
\centering
\small
\begin{tabular}{lrrrr}
\toprule
Objective & F-SWD$\downarrow$ & Energy$\downarrow$ & Precision$\uparrow$ & Recall$\uparrow$\\
\midrule
Rectified flow & .904 & 2.929 & .335 & .541\\
GFT diffusion, noise prediction & 4.049 & 38.389 & .000 & .000\\
GFT diffusion, velocity prediction & 1.151 & 5.070 & .159 & .429\\
\bottomrule
\end{tabular}
\end{table}

\section{Condition-Matched DiffeoCFM Comparison}
\label{sec:matchedcategorical}
This additional experiment compares CorrFlow with a pinned adaptation of the
official DiffeoCFM correlation implementation \citep{collas2025diffeocfm}, its
full class-wise OAS Gaussian prior, historical replay and a categorical cCorrGAN
reconstruction. All receive only the
categorical regime, on the same return-derived Ward-ordered Taiwan matrices.
The original regime-plus-continuous-score experiment remains a separate table.
The adapted DiffeoCFM transform, field, objective and gradients were checked
against the pinned implementation; this is more than a coordinate surrogate.

CorrFlow uses its 48-channel, three-block field for 20 epochs; DiffeoCFM uses
the 512-wide float64 SELU field for 200; cCorrGAN also trains for 200.
These are declared method-specific
budgets, not equal compute or exhaustive tuning. Approximately matched numbers
of validation comparisons select minimum full structural SWD on the same 300
validation cases (100 per regime), with fixed latent draws. Five training seeds
pair with five generation seeds. All methods use the same 1,500 test cases,
500 per regime; controls have five generation seeds from one fitted dataset.
The test periods were inspected during development.

CorrFlow sampling uses 25 Heun steps (50 field calls) and the converged float64
GFT inverse; DiffeoCFM uses adaptive Dopri5 with absolute/relative tolerances
\(10^{-5}\). Common scoring uses float32 matrices. A provenance audit verifies
hashes of code, data and output artifacts, completed budgets, identical case
arrays, replay membership and validation selection. The ten native selection
scores, together with the five cCorrGAN scores, reproduce exactly. All evaluated
samples are valid after required image-model repair; this does not establish
raw cCorrGAN validity.

\begin{table}[t]
\caption{Matched categorical Taiwan comparison. Identical 1,500 test cases, 500 per regime. Means and sample standard deviations over five training/paired generation seeds for neural models; five draws from the same fitted training data for controls. Selected weights; CorrFlow 20 epochs, DiffeoCFM 200 and cCorrGAN 200, not equal compute.}
\label{tab:matchedcategorical}
\centering\small
\setlength{\tabcolsep}{5.5pt}
\begin{tabular}{lrrrrr}
\toprule
Method & F-SWD $\downarrow$ & Precision & Recall & Portfolio SWD $\downarrow$ & C2ST AUC \\
\midrule
CorrFlow & \(0.471\pm0.062\) & 0.524 & 0.320 & \(0.624\pm0.094\) & 0.999 \\
DiffeoCFM & \(0.751\pm0.019\) & 0.304 & 0.355 & \(0.862\pm0.026\) & 0.997 \\
Transformed Gaussian & \(0.871\pm0.006\) & 0.301 & 0.504 & \(0.811\pm0.013\) & 0.995 \\
Training replay & \(0.877\pm0.008\) & 0.450 & 0.590 & \(1.107\pm0.009\) & 0.955 \\
cCorrGAN reconstruction & \(4.039\pm0.154\) & 0.000 & 0.401 & \(818.722\pm824.701\) & 1.000 \\
\bottomrule
\end{tabular}
\end{table}

CorrFlow has lower structural and portfolio distances under this protocol
(Table~\ref{tab:matchedcategorical}). DiffeoCFM has slightly higher recall,
and replay has higher recall than both. DiffeoCFM improves structural distance
relative to its Gaussian prior, but worsens portfolio distance and reduces
recall. Both learned models remain readily distinguishable from real data:
their two-sample classifier AUCs exceed .996. These are complementary findings,
not evidence of universal superiority or complete distribution recovery.

Checkpoint selection matters. The selected CorrFlow epochs are 1, 3, 18, 6 and
4; DiffeoCFM selects 170, 190, 140, 200 and 130. Final-weight structural SWD is
\(.836\pm.068\) for CorrFlow and \(.798\pm.022\) for DiffeoCFM, reversing
the selected-checkpoint ordering. We retain this declared sensitivity without
replacing the predeclared validation-selection rule after seeing test results.

GFT distance computed in Ward order favors DiffeoCFM (.108 versus .335).
Distance after independent random relabeling favors CorrFlow (.077 versus
.086). These metrics answer different questions about the coordinate order.
The full five-seed cCorrGAN reconstruction has structural SWD
\(4.039\pm.154\), zero measured precision and portfolio SWD
\(818.722\pm824.701\) after repair. Its poor fidelity and large portfolio
dispersion remain part of the comparison; completing its budget does not
reproduce the original publication's training fidelity.
Seed variability does not quantify uncertainty over independent market
episodes. Concurrent runtimes are unsuitable for speed rankings.

\section{External Japan Diagnostic}
\label{app:japan}

A separate archived experiment uses the JPX stock-prediction dataset, with
3,848 training matrices (2018--2019), 1,936 validation matrices (2020), and
2,888 test matrices (2021--2022). Each matrix again uses 80 stocks and 252
sessions. Returns are adjusted for splits but not cash dividends, unlike the
Taiwan exchange-reference return panel. The public archive ends in June 2022, so this is an
external market rather than a later temporal holdout. These experiments were
completed during development and are reported here as supplementary evidence.
An audit of the actual 252-session windows finds 251 shared return sessions at
each adjacent split boundary and nine shared sessions between the original
training and test windows. Japan's 242-session validation endpoint period is
too short to purge the first 251 endpoints while retaining any validation
matrix. This diagnostic is therefore not evidence from observation-disjoint
splits; a future benchmark requires redesigned dates and new fits.

CorrFlow and velocity-prediction GFT diffusion use the same architecture,
20-epoch budget, five seeds, raw minimum-validation-loss weights, and 50 network
evaluations. Evaluation uses 180 cases (60 per regime), limited by the 67
adverse-regime examples in the test split. All neural inverses converge and all
samples are valid. Table~\ref{tab:japan} shows that CorrFlow's structural-distance
advantage over this matched diffusion configuration persists, but neither
neural generator beats historical replay. Thus the Taiwan ranking against
replay does not generalize automatically to another market.

\begin{table}[h]
\caption{Japan diagnostic. Neural distances are means (sample s.d.) over five
training seeds; precision and recall show seed means. Controls use one generation seed. Different sample counts and
market-specific feature scales preclude direct comparison of distances with
the Taiwan table.}
\label{tab:japan}
\centering
\small
\begin{tabular}{lrrrr}
\toprule
Model & F-SWD$\downarrow$ & Energy$\downarrow$ & Precision$\uparrow$ & Recall$\uparrow$\\
\midrule
Historical replay & 1.013 & 2.922 & .550 & .144\\
PCA-64 Gaussian & 1.725 & 9.020 & .006 & .000\\
CorrFlow & 1.374 (.114) & 5.480 (.860) & .036 & .017\\
GFT diffusion, velocity & 1.641 (.116) & 7.525 (.775) & .023 & .008\\
\bottomrule
\end{tabular}
\end{table}

A 1,000-repeat bootstrap jointly resampling the 136 window-end dates and paired
training seeds gives a 95\% percentile interval of [.148, .373] for the
F-SWD difference (diffusion minus flow) and [1.108, 2.829] for energy distance.
Grouping by date accounts for same-date asset subsets but not all serial
dependence between overlapping 252-session windows. These intervals therefore
do not establish uncertainty across independent market regimes.

\section{Computational Cost and Size Limits}
\label{app:runtime}

The $O(n^2)$ network cost assumes fixed hidden width and depth. Each GFT
inverse iteration performs a dense symmetric eigendecomposition to exponentiate
the completed log matrix, costing $O(n^3)$; iteration count also depends on the
coordinates and tolerance. Higham repair similarly requires repeated cubic-cost
spectral operations. It is therefore inappropriate to compare CorrFlow's full
sampler with only the forward pass of an image generator.

Archived elapsed times for 20 training epochs on the shared GB10 are
$919\pm49$ seconds for CorrFlow, $593\pm92$ for cCorrGAN, and $586\pm71$ for
image DDPM (five seeds; includes the training loop and validation). These are
observed run costs, not isolated-device throughput benchmarks. CorrFlow's
recorded generation time is $60.8\pm4.8$ seconds per 1,000 matrices before
inverse GFT. Inversion time is reported separately in the archived evaluation
records. The image runs did not record comparable sampling-plus-repair times,
so no end-to-end speed ranking follows from these logs.

\textbf{Synthetic scaling benchmark.} A subsequent GB10 benchmark measures the
48-channel, three-block network with batch size one and bfloat16 autocasting.
Weights are initially random, with a warm-up update followed by three timed
training updates at each size. ODE timings use 25 Heun steps; all GPU timings
are synchronized medians of three repetitions after warm-up. Inverse timings
are separate: they invert synthetic Gaussian-factor sample correlations based
on $\lceil3.2n\rceil$ observations, not the ODE endpoints. Thus these measurements
characterize computation rather than learned financial fidelity or complete
training cost. Every synthetic inverse converged at tolerance $10^{-6}$.

\begin{table}[h]
\caption{Synthetic GB10 scaling at batch size one. Memory is peak PyTorch
allocated memory during training, excluding other processes. The two sampling
stages use different synthetic inputs and are not an end-to-end latency measurement.}
\label{tab:scaling}
\centering
\small
\begin{tabular}{rrrrr}
\toprule
Assets & Training step (s) & Peak training (GiB) & ODE (s/matrix) & Inverse GFT (s/matrix)\\
\midrule
80 & .006 & .097 & .042 & .009\\
160 & .009 & .192 & .144 & .008\\
200 & .014 & .260 & .238 & .012\\
500 & .101 & 1.282 & 1.867 & .028\\
1,000 & .389 & 4.933 & 7.553 & .082\\
\bottomrule
\end{tabular}
\end{table}

At 1,000 assets, peak training memory is 4.93 GiB and a training update takes
0.389 s at batch size one. These measurements demonstrate computational
feasibility at the thousand-asset scale. The architecture and GFT construction
support larger dimensions; computational cost grows with matrix size, and
inverse iterations also depend on matrix conditioning.

\subsection{Statistical, computational, and learned size limits}
\label{app:size}

With $T$ complete observations and sample-mean removal,
$\operatorname{rank}(\widehat C)\leq\min(n,T-1)$. Thus $T=252$ permits full
rank only for $n\leq251$; even below this ceiling, severe ill-conditioning can
make estimates unreliable. Dimensions above it require longer histories or an
explicitly regularized estimator and consequently a revised empirical target.
The GFT bijection itself has no fixed dimension ceiling. Longer available
US and Taiwan histories permit larger $T$, but preserving $n/T\approx80/252$
would require about 1,575 sessions at $n=500$ and 3,150 at $n=1{,}000$.
These longer windows mix more market regimes and require sufficient complete
asset histories; they are study designs rather than demonstrated results.

As a separate data-conditioning diagnostic, the archived Japan benchmark
contains exactly nested 20/40/80/160/200-asset estimates from 252 sessions.
Median training condition numbers rise from 306 at $n=80$ to 2,746 at $n=160$
and 11,550 at $n=200$; validation reaches 38,676 at $n=200$. All are positive
definite. These figures describe an estimator stress test, not learned
CorrFlow fidelity at those sizes or a hardware capacity benchmark.

Parameter sharing permits execution at different sizes, but learned
cross-size accuracy is not automatic. Applying the 80-asset checkpoint directly
at $n=40$ gave F-SWD 2.070 and precision/recall .128/.245 in a single-seed
diagnostic, despite 100\% validity. For the same checkpoint, direct 40-asset
draws also differed from 40-asset principal submatrices of 80-asset draws
(F-SWD 1.332). Evaluation at a new dimension should include training and
generation fidelity, peak memory, vector-field runtime, inverse convergence,
and total sampling time.

\section{Temporal Spectral Shift}
\label{app:shift}

\begin{figure}[!hbp]
\centering
\includegraphics[width=.98\textwidth]{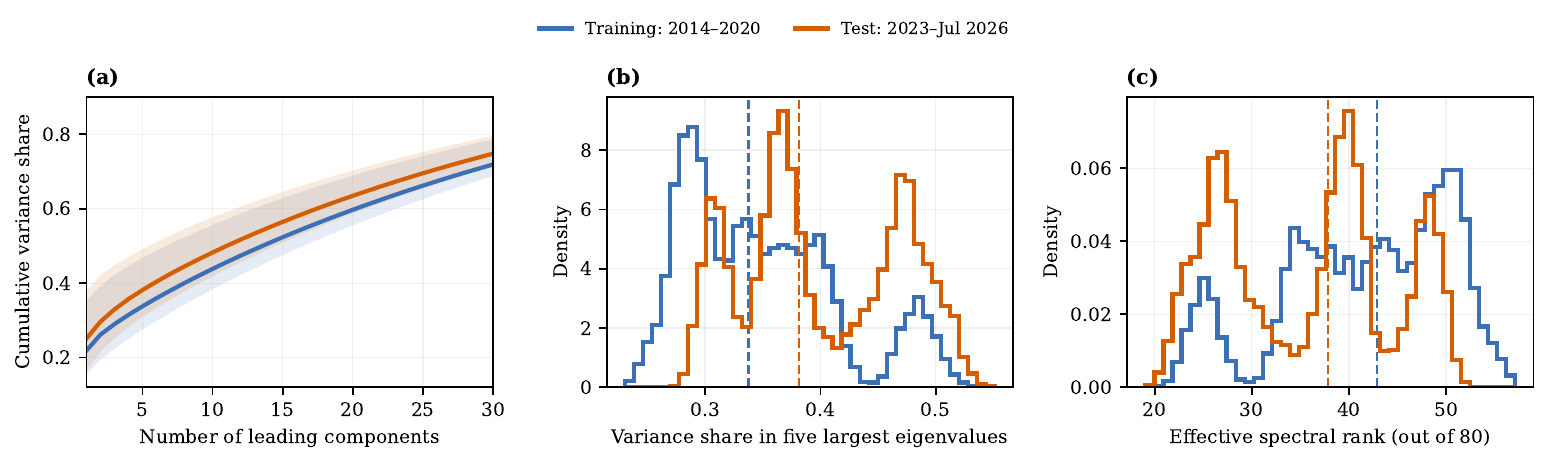}
\caption{The spectral change between the full, naturally weighted training and
test splits. (a) Median cumulative variance explained by the leading
eigenvalues; bands span the 10th--90th percentiles across matrices. (b) Test
matrices shift toward a larger share in the five leading eigenvalues (median
.382 versus .338). (c) Their entropy-based effective rank shifts lower (median
37.9 versus 42.9 out of 80). Dashed lines mark medians. The test period therefore contains
fewer effective directions of standardized variation, consistent with stronger
common movement and less diversification potential for broad portfolios.}
\label{fig:spectral-shift}
\end{figure}

Figure~\ref{fig:spectral-shift} compares the eigenvalue spectra of all Taiwan
training and test matrices. Each matrix receives equal weight, preserving the
observed regime frequencies within each split. Table~\ref{tab:final} evaluates
a balanced subset of 1,500 test cases, with 500 from each regime.

\end{document}